# Learning Protein Dynamics with Metastable Switching Systems

Bharath Ramsundar[1] and Vijay S. Pande[2]
[1] Department of Computer Science, Stanford University, Stanford CA 94305, USA
[2] Department of Chemistry, Stanford University, Stanford CA 94305, USA

(Dated: 5 October 2016)

We introduce a machine learning approach for extracting fine-grained representations of protein evolution from molecular dynamics datasets. Metastable switching linear dynamical systems extend standard switching models with a physically-inspired stability constraint. This constraint enables the learning of nuanced representations of protein dynamics that closely match physical reality. We derive an EM algorithm for learning, where the E-step extends the forward-backward algorithm for HMMs and the M-step requires the solution of large biconvex optimization problems. We construct an approximate semidefinite program solver based on the Frank-Wolfe algorithm and use it to solve the M-step. We apply our EM algorithm to learn accurate dynamics from large simulation datasets for the opioid peptide met-enkephalin and the proto-oncogene Src-kinase. Our learned models demonstrate significant improvements in temporal coherence over HMMs and standard switching models for met-enkephalin, and sample transition paths (possibly useful in rational drug design) for Src-kinase.

## I. INTRODUCTION

Understanding protein folding and conformational change is a central challenge in modern biology. Since proteins are microscopic objects, experimental techniques cannot satisfactorily resolve protein dynamics due to limited spatial and temporal resolution. Molecular dynamics (MD) simulations complement experimental understanding by providing a computational lens into the high-frequency atomic dynamics of proteins[1]. MD simulations forward-integrate Newton's equations of motion for classical approximations of the energy landscapes of biological macromolecules[2]. These simulations have thousands of degrees of freedom which must be integrated over billions of timesteps to sample biologically relevant events[3]. Machine learning techniques which condense MD datasets into human-comprehensible representations can provide crucial insight into protein structure and dynamics.

Several machine-learning methods have been proposed for analyzing protein trajectories. Dimensionality reduction methods such as PCA provide insight into the high-variance degrees of freedom of the system[4], but do not model the temporal structure of the data. More sophisticated methods such as Markov state models (MSMs)[5] and hidden Markov models (HMMs)[6] identify long-lived (metastable) conformations of the protein and model dynamics by a Markovian jump process between metastable states. These methods represent the state-of-the-art for analyzing MD trajectories and have been widely adopted[7].

Despite their strengths, HMMs and MSMs only parameterize discrete jump processes between metastable states and do not describe the complex, continuous dynamics of the underlying physical system. Consequently, these systems cannot accurately model transition paths between metastable conformations. This limitation is unfortunate, since detailed understanding of the physical transitions between metastable states could facilitate the rational design of targeted drugs[8].

Parameterizing the dynamics of a protein requires modeling the underlying physical Hamiltonian. Such modeling is complicated by the high-dimensional and intrinsically nonconvex potential energy surface of the system. HMMs and MSMs can identify the local minima of this energy landscape but make no effort to describe the interstitial terrain. Prior work has suggested using locally quadratic approximations to the Hamiltonian, which have the convenient property that the emergent dynamics are locally described by linear operators[9]. The associated probabilistic model is the switching linear dynamical system[10].

Standard switching models are not well suited for modelling physical systems, since the associated linear operators must be constrained to prevent degeneracy as time goes to infinity. Earlier work has suggested constraining the spectral radius of the operators to have magnitude at most $1$[11]. While this constraint is sufficient given negligible system noise, physical systems with stochastic dynamics require stronger constraints to prevent degeneracy. The proper constraint can be derived from classical notions of Lyapunov stability in control theory[11]. Let $\Sigma_s$ be a fixed upper bound on the desired covariance for metastable state $s$. Let $A_s$ be a linear model for system evolution in state $s$ and let $Q_s$ model Brownian noise. The constraint $Q_s + A_s \Sigma_s A_s^T \preceq \Sigma_s$ (that is, $\Sigma_s - Q_s - A_s \Sigma_s A_s^T$ is a positive semidefinite matrix) guarantees that the learned model for metastable state $s$ will have covariance upper bounded by $\Sigma_s$.

We define a metastable switching linear dynamical system to be a switching model with covariance constraint $Q_s + A_s \Sigma_s A_s^T \preceq \Sigma_s$ and spectral norm constraint $\|A_s\|_2 < 1$ for each metastable state $s$. Section II introduces the graphical model associated with a metastable switching system and proves the sufficiency of our stability constraints. Section III establishes a learn-ing procedure for metastable switching models based on expectation-maximization. Section III A derives the E-step by adapting the forward-backward procedure for HMMs, and section III B reduces the M-step to the solution of a constrained biconvex optimization problem. This problem is convex in each variable with semidefinite cone constraints. The interior point method can solve such programs[12], but constructs inverse Hessian matrices at a cost of $O(d^6)$ per iteration for $d$ degrees of freedom. Protein systems typically have hundreds to thousands of interesting degrees of freedom, too many for efficient solution via the interior point method. Section IV generalizes prior work on Frank-Wolfe algorithms[13] to learn $A_s$ and $Q_s$ at a cost of $O(d^2)$ per iteration. Section V learns metastable switching models for the opioid peptide met-enkephalin[14] and the proto-oncogene Src-Kinase[15] and demonstrates that the emergent dynamics are significantly more nuanced than those learned by HMMs and standard switching systems. The learned model for Src-Kinase succeeds in sampling transition paths between metastable conformations; such paths may be useful for the rational design of targeted kinase inhibitors for cancer treatments[15].

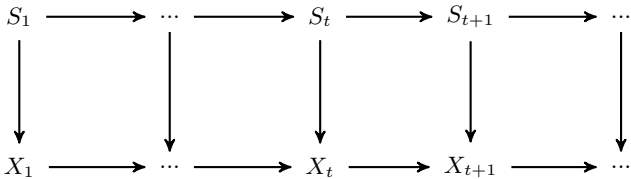

FIG. 1: Graphical model for metastable switching linear dynamical systems

## II. METASTABLE SWITCHING LINEAR DYNAMICAL SYSTEMS

Let $\{X_t\}$ be an observed time series in $\mathbb{R}^D$, and let $\{S_t\}$ be a hidden time series of latent states in $\{1, \ldots, K\}$. For molecular dynamics trajectories, each hidden state corresponds to a metastable state of the physical system, while each observed state corresponds to a physical conformation of the protein. A metastable switching linear dynamical system is a generative probabilistic model over the $S_t$ and $X_t$. The graphical model representation of this model is shown in Figure 1.

The hidden states $S_t$ evolve in a Markovian manner according to probability matrix $\mathbf{T} \in \mathbb{R}^{K \times K}$, where $\mathbf{T}_{s_1, s_2}$ is the probability of hidden state $s_1$ transitioning to hidden state $s_2$. The observed states $X_t$ evolve by an affine transformation of $X_{t-1}$ parameterized by hidden state $s_t$ (we follow that convention that upper case $X_t$ denotes the random variable and lower case $x_t$ a particular value):

$$x_t = A_{s_t} x_{t-1} + b_{s_t} + w_{s_t}$$

The linear operator $A_{s_t}$ and affine shift vector $b_{s_t}$ control the evolution of the system in metastable state $s_t$. The terms $w_{s_t}$ are noise parameters drawn from distribution $\mathcal{N}(0, Q_{s_t})$, where $Q_{s_t}$ is a "local" covariance matrix that corresponds to Brownian noise. To guarantee that the model is stable, iterating the transformation $A_{s_t} x_{t-1} + b_{s_t} + w_{s_t}$ must not lead to unbounded mean or covariance.

Let the initial state be $x_0$ and the metastable state $s$. The distribution of $X_1$ is $\mathcal{N}(A_s x_0 + b_s, Q_s)$, while the distribution of $X_2$ is $\mathcal{N}(A_s^2 x_0 + A_s b_s + b_s, Q_s + A Q_s A^T)$. Iterating, the distribution of $X_n$ is

$$\mathcal{N}\left(A_s^n x_0 + \sum_{k=0}^{n-1} A_s^k b, \sum_{k=0}^{n-1} A_s^k Q_s (A_s^k)^T\right).$$

To achieve an accurate model, this distribution must match the true distribution of metastable state $s$. Let $\mathcal{N}(\mu_s, \Sigma_s)$ be a Gaussian model of the distribution of physical conformations associated with metastable state $s$. If $b_s = \mu_s - A_s \mu_s$, the mean of the iterated distribution above converges to $\mu_s$ if the spectral norm of $A_s$ is less than 1.

**Lemma 1.** *If $\|A_s\|_2 < 1$, and $b_s = \mu_s - A_s \mu_s$, then $A_s^n x_0 + \sum_{k=0}^{n-1} A_s^k b$ converges to $\mu_s$ as $n \to \infty$.*

*Proof.* Since $\|A_s\|_2 < 1$, the summation $\sum_{k=1}^{\infty} A_s^k$ converges to $(I - A_s)^{-1}$ and the limit $\lim_{n \to \infty} A^n$ equals 0. Thus, $\lim_{n \to \infty} \left(A_s^n x_0 + \sum_{k=0}^{n-1} A_s^k b\right)$ equals $(I - A_s)^{-1} b_s$, which in turn equals $\mu_s$ if and only if $b_s = \mu_s - A_s \mu_s$. □

The next lemma proves that enforcing constraint $Q_s + A_s \Sigma_s A_s^T \preceq \Sigma_s$ guarantees that the iterated distribution has covariance upper bounded by $\Sigma_s$ (recall that the constraint $A \preceq B$ denotes that $B - A$ is a positive semidefinite matrix).

**Lemma 2.** *Assume that $Q_s + A_s \Sigma_s A_s^T \preceq \Sigma_s$ and that $Q_s \preceq \Sigma_s$. Then for all $n \in \mathbb{N}$, $\sum_{k=0}^{n-1} A_s^k Q_s (A_s^k)^T \preceq \Sigma_s$.*

*Proof.* Since $Q_s \preceq \Sigma_s$, it follows that $Q_s + A_s Q_s A_s^T \preceq Q_s + A_s \Sigma_s A_s^T \preceq \Sigma_s$. The result follows by induction. □

## III. LEARNING

A metastable switching model is parameterized by choice of linear operators $\{A_s\}$, affine shifts $\{b_s\}$, local covariances $\{Q_s\}$, and transition matrix $\mathbf{T}$. The complete-data likelihood for the model is

$$\mathcal{L}(\{S_t\}, \{X_t\} \mid T, \{A_s\}, \{b_s\}, \{Q_s\}) = \prod_{t=1}^{T} \mathbf{T}_{s_{t-1}, s_t} \prod_{t=1}^{T} \mathcal{N}(x_t \mid A_{s_t} x_{t-1} + b_{s_t}, Q_{s_t})$$

We use expectation-maximization to learn the model parameters by optimizing the log-likelihood of observed data $\{X_t\}$. Section III A describes the inference procedure for the E-step, and section III B reduces the M-step to the solution of a biconvex optimization problem.

## A. E-step

The E-step for a metastable system estimates hidden states $\{S_t\}$ given model parameters $(\mathbf{T}, \{A_s\}, \{b_s\}, \{Q_s\})$ and observed data $\{X_t\}$ by using a variant of the forward-backward algorithm for Gaussian HMMs[16]. We derive the recursive forward updates that compute $\alpha(s_t) = \mathbb{P}(s_t \mid x_{1:t})$, the posterior on hidden state $S_t$ given observed data $x_{1:t}$.

$$\begin{aligned}\alpha(s_{t+1}) &= \mathbb{P}(s_{t+1}|x_{1:t+1}) \\ &\propto \mathbb{P}(x_{t+1}|s_{t+1}, x_t) \sum_{s_t} \mathbb{P}(s_{t+1}|s_t)\mathbb{P}(s_t|x_{1:t}) \\ &= \mathcal{N}(x_{t+1}|A_{s_{t+1}}x_t + b_{s_{t+1}}, Q_{s_{t+1}}) \sum_{s_t} \mathbf{T}_{s_t, s_{t+1}} \alpha(s_t)\end{aligned}$$

We analogously derive the backward updates for $\beta(s_t) = \mathbb{P}(x_{t+1:T} \mid s_t, x_t)$, the probability of the future data conditioned on $s_t$ and $x_t$.

$$\begin{aligned}\beta(s_t) &= \mathbb{P}(x_{t+1:T} \mid s_t, x_t) \\ &= \sum_{s_{t+1}} \mathbb{P}(x_{t+2:T}|s_{t+1}, x_{t+1})\mathbb{P}(x_{t+1} \mid s_{t+1}, x_t) \\ &\quad \times \mathbb{P}(s_{t+1}|s_t, x_t) \\ &= \sum_{s_{t+1}} \beta(s_{t+1}) \mathcal{N}(x_{t+1} \mid A_{s_{t+1}}x_t + b_{s_{t+1}}, Q_{s_{t+1}}) \mathbf{T}_{s_t, s_{t+1}}.\end{aligned}$$

In both derivations, we use the fact that $d$-separation implies that

$$\mathbb{P}(s_{t+1}|s_t, x_t) = \mathbb{P}(s_{t+1}|s_t) = \mathbf{T}_{s_t, s_{t+1}}.$$

We compute the posterior probability on hidden states $\gamma_{s_t}(t)$ as for a HMM.

$$\gamma_{s_t}(t) = \mathbb{P}(s_t \mid x_{1:T}) \propto \beta(s_t)\alpha(s_t)$$

Let $x'_{t+1} = A_{s_{t+1}}x_t + b_{s_{t+1}}$. Then the joint transition probability $\xi_{s_t, s_{t+1}}$ is

$$\begin{aligned}\xi_{s_t, s_{t+1}} &= \mathbb{P}(s_t, s_{t+1} \mid x_{1:T}) \\ &= \frac{\alpha(s_t) T_{s_t, s_{t+1}} \beta(s_{t+1}) \mathcal{N}(x_{t+1} \mid x'_{t+1}, Q_{s_{t+1}})}{\sum_s \alpha(s)\beta(s)}\end{aligned}$$

### 1. Implementation

The E-step is implemented as a multithreaded CPU application. The quantities $\alpha, \beta$ are computed in parallel across MD trajectories during the E-step using OpenMP. The largest portion of the run time is spent in LOG-SUM-EXP operations, which are manually vectorized with SSE2 intrinsics for SIMD architectures.

## B. M-step

The M-step selects model parameters $(\mathbf{T}, \{A_s\}, \{Q_s\}, \{b_s\})$ that optimize the complete-data log-likelihood for the model. We assume that for each metastable state $s$, we are given a Gaussian $\mathcal{N}(\mu_s, \Sigma_s)$ fitted to the data from this state. The mean $\mu_s$ and $\Sigma_s$ are used to constrain the optimization problem to ensure stability, but these quantities are not themselves estimated in the M-step; they should be estimated in a preliminary step by fitting a Gaussian mixture model to the data.

### 1. Learning T

The transition matrix $\mathbf{T}$ is learned as for the hidden Markov model:

$$T_{i,j} = \frac{\sum_t \xi_{i,j}(t)}{\sum_t \gamma_i(t)}$$

### 2. Objective for a fixed metastable state $s$

In principle, the terms $(\{A_s\}, \{Q_s\}, \{b_s\})$ should be chosen to maximize the log-likelihood of the observed data. Let $x'_{t+1} = A_{s_t} x_{t-1} + b_{s_t}$.

$$\log \mathcal{L}(\{X_t\} \mid T, \{A_s\}, \{b_s\}, \{Q_s\}) = \\ \int \sum_{t=1}^{T} \left[ \mathbf{T}_{s_{t-1}, s_t} + \sum_{t=1}^{T} \log \mathcal{N}(x_t \mid x'_{t+1}, Q_{s_t}) \right] dS_{1:T}.$$

The objective above is intractable, so we instead exploit an approximation of the objective for $(A_s, Q_s, b_s)$ that weights the terms for time $t$ with posterior likelihood $\gamma_s(t)$[10]. With the stability constraints added, the joint optimization problem becomes

$$\begin{aligned}\underset{A_s, Q_s, b_s}{\text{minimize}} \quad & \log \det(Q_s) \left( \sum_t \gamma_s(t) \right) \\ & + \sum_t \gamma_s(t)(x_t - A_s x_{t-1} - b)^T Q_s^{-1} (x_t - A_s x_{t-1} - b_s) \\ \text{subject to} \quad & Q_s + A_s \Sigma_s A_i^T \preceq \Sigma_s, \quad \|A_s\|_2 < 1 \\ & A_s \mu_s + b_s = \mu_s.\end{aligned}$$

Note that constraint $A_s \mu_s + b_s = \mu_s$ uniquely determines $b_s$. Thus, the optimization above depends only on $A_s$ and $Q_s$. We show in the following sections that this objective is convex in both variables.

### 3. Learning $A_s$

Consider the objective with $Q_s$ and $b_s$ fixed. The $\log \det$ can be dropped, and the cyclic invariance of the trace can be used to factor out $Q_s^{-1}$:

$$\text{Tr}\left( Q_s^{-1} \sum_t \gamma_s(t)(x_t - A_s x_{t-1} - b_s)(x_t - A_s x_{t-1} - b_s)^T \right).$$

The summation can be simplified by dropping terms not dependent on $A_s$:

$$\operatorname{Tr}\left(Q_s^{-1}\left[-B_s A_s^T - A_s B_s^T + A_s E_s A_s^T + A_s C_s^T + C_s A_s^T\right]\right).$$

In this expression, we use sufficient statistics matrices calculated from the data in the E-step:

$$C_s = b \sum_t \gamma_s(t) x_{t-1}^T$$

$$B_s = \sum_t \gamma_s(t) x_t x_{t-1}^T$$

$$E_s = \sum_t \gamma_s(t) x_{t-1} x_{t-1}^T$$

Consider the constraint $Q_s + A_s \Sigma_s A_s^T \preceq \Sigma_s$. This expression can be rewritten by using the Schur complement; since the covariance upper bound $\Sigma_s$ is positive definite, the matrix

$$\begin{pmatrix} \Sigma_s - Q_s & A_s \\ A_s^T & \Sigma_s^{-1} \end{pmatrix}$$

is positive semidefinite if and only if $Q_s + A_s \Sigma_s A_s^T \preceq \Sigma_s$. Similarly, the constraint $\|A_s\|_2 < 1$ can be rewritten as a Schur complement; the following matrix is positive semidefinite if and only if $\|A_s\|_2 \leq \eta$, where $\eta < 1$ is a fixed constant:

$$\begin{pmatrix} \eta I & A_s \\ A_s^T & I \end{pmatrix} \succeq 0.$$

Now, the optimization problem for $A_s$ is a quadratic matrix program with two linear matrix matrix inequalities. Let $F_s = C_s - B_s$. Then

$$\begin{aligned}
\underset{A_s}{\text{minimize}} \quad & \operatorname{Tr} Q_s^{-1}\left(F_s A_s^T + A_s F_s^T + A_s E_s A_s^T\right) \\
\text{subject to} \quad & \begin{pmatrix} \Sigma_s - Q_s & A_s \\ A_s^T & \Sigma_s^{-1} \end{pmatrix} \succeq 0, \quad \begin{pmatrix} \eta & A_s \\ A_s^T & I \end{pmatrix} \succeq 0.
\end{aligned}$$

#### 4. Learning $Q_s$

Consider the objective with $A_s$ and $b_s$ fixed. Applying the cyclic invariance of the trace simplifies the objective to

$$\log \det(Q_s) g_s + \operatorname{Tr}\left(Q_s^{-1} F_s\right),$$

where $F_s$ and $g_s$ are sufficient statistics matrices computed from the data:

$$F_s = \sum_t \gamma_s(t)(x_t - A_s x_{t-1} - b_s)(x_t - A_s x_{t-1} - b_s)^T$$

$$g_s = \left(\sum_t \gamma_s(t)\right).$$

We simplify the objective further by applying the change of variable $R_s = Q_s^{-1}$:

$$-\log \det R_s g_s + \operatorname{Tr}(R_s F_s).$$

The stability constraint becomes $R_s^{-1} + A_s \Sigma_s A_s^T \preceq \Sigma_s$. Applying the Schur complement as before, the optimization problem for $Q$ becomes the following semidefinite program:

$$\begin{aligned}
\underset{R_s}{\text{minimize}} \quad & -\log \det(R_s) g_s + \operatorname{Tr}(R_s F_s) \\
\text{subject to} \quad & \begin{pmatrix} \Sigma_s - A_s \Sigma_s A_s^T & I_n \\ I_n & R_s \end{pmatrix} \succeq 0.
\end{aligned}$$

## IV. SOLVING FOR $\{A_s\}$, $\{Q_s\}$ WITH A FRANK-WOLFE ALGORITHM

The learning problems for $A_s$ and $Q_s$ are convex programs with semidefinite cone constraints. Although interior point solvers routinely solve small cone programs[12], these methods require the construction of large Hessian and inverse Hessian matrices. Thus, we use a first-order Frank-Wolfe algorithm[17] to solve for $A_s$ and $Q_s$ in large protein systems. We extend prior work that uses a Frank-Wolfe method to solve semidefinite programs (SDPs)[13].

### A. Generalized penalty

The Frank-Wolfe SDP solver encodes linear matrix inequality constraints within a penalty function. The minima of the penalty function lies below some threshold if and only if the SDP is approximately feasible. A binary search then enables the solution of general SDPs with logarithmic overhead. This algorithm cannot solve for $A_s$ and $Q_s$, since the objective functions are nonlinear. We introduce a new penalty function which resolves this issue. Let $f_1, \ldots, f_N$ be convex functions on $\mathbb{R}^{N \times N}$ with associated inequality constraints $f_i(X) \leq 0$, and let $g_j(X)$ be affine functions on $\mathbb{R}^{N \times N}$ with associated equality constraints $g_j(X) = 0$. Define a log-sum-exp penalty function

$$\Phi[\{f_i\}, \{g_j\}](X) =$$

$$\frac{1}{M} \log \left( \sum_{i=1}^n \exp(M f_i(X)) + \sum_{j=1}^m \exp\left(M g_j(X)^2\right) \right)$$

Note that $\Phi[\{f_i\}, \{g_j\}]$ (or $\Phi$ for short) is a convex function by the rules of convex composition[18], since the $f_i$ and $g_j^2$ are convex and the log-sum-exp function is convex and nondecreasing in each argument. Penalty $\Phi$ lies below some threshold $\epsilon$ if and only if $X$ approximately satisfies constraints $f_i(X) \leq 0$ and $g_j(X) = 0$.

**Lemma 3.** *Let $M = \log(n+m)/\epsilon$. For all $X \in \mathbb{R}^{N \times N}$, if objective $\Phi(X)$ is upper bounded by $\epsilon$, then $X$ satisfies* constraints $f_i(X) \leq \epsilon$ and $g_j(X)^2 \leq \epsilon$. Conversely, if $X$ is such that $f_i(X) \leq \epsilon$ and $g_j(X)^2 \leq \epsilon$ for all $i$ and $j$, then $\Phi(X) \leq 2\epsilon$.

*Proof.* Define $\phi(x) = \frac{1}{M} \log \sum_{i=1}^{\ell} \exp(Mx_i)$. For $M \geq 0$, the function $\phi$ is a smooth approximation to the max function[13]:

$$\max_i x_i \leq \phi(x) \leq \max_i x_i + \frac{\log \ell}{M}$$

Thus if $\Phi(X) \leq \epsilon$, then $X$ satisfies condition $\max_{i,j}\{f_i(X), g_j(X)^2\} \leq \Phi(X) \leq \epsilon$. Conversely, suppose $X$ is such that $f_i(X) \leq \epsilon$ and $g_j^2(X) \leq \epsilon$. The upper bound above, combined with choice of $M$, guarantees that $\Phi(X) \leq 2\epsilon$. □

### B. Feasibility Search

---
**Algorithm 1** Feasibility search
---
1: **procedure** FEASIBILITY-SEARCH($h, \{f_i\}, \{g_j\}, \epsilon, R, K$)
2: ▷ $h$ objective, $f_i(X) \leq 0, g_j(X) = 0$ constraints, $\epsilon > 0$ error threshold
3:     $X \leftarrow$ Frank-Wolfe($\Phi[\{f_i\}, \{g_j\}], R, K$)
4:     **if** $\Phi[\{f_i\}, \{g_j\}](X) < \epsilon$ **then**
5:         $U \leftarrow h(X), \eta \leftarrow 1$ ▷ $\eta$ is step size
6:         **while** $\eta \geq \epsilon$ **do**
7:             $h_{U-\eta} \leftarrow \lambda Y : h(Y) - (U - \eta)$
8:             $X_{U-\eta} \leftarrow$ Frank-Wolfe($\Phi[\{h_{U-\eta}, f_i\}, \{g_j\}], R, K$)
9:             **if** $\Phi[\{h_{U-\eta}, f_i\}, \{g_j\}](X_{U-\eta}) < \epsilon$ **then**
10:                 $X \leftarrow X_{U-\eta}, U \leftarrow h(X), \eta \leftarrow 2\eta$
11:             **else**
12:                 $\eta \leftarrow \frac{1}{2}\eta$
13:         **return** Success, $X$
14:     **return** Fail
---

---
**Algorithm 2** Frank-Wolfe
---
**procedure** FRANK-WOLFE($\Phi, R, K$)
▷ $R$ upper bound on tr($X$), $K$ number of desired descent steps
    $X \leftarrow$ RESCALE($X$)
    **for** $k \leftarrow 1 \ldots K$ **do**
        $v_k \leftarrow$ APPROX-EV($-\nabla\Phi(X)$)
        Line search for $\gamma$
        $X \leftarrow (1-\gamma)X + \gamma v_k v_k^T$
    **return** $X$
---

We introduce a feasibility search method in algorithm 1 which optimizes a convex objective $h$ while attempting to satisfy constraints $f_i(X) \leq 0$ and $g_j(X) = 0$. The procedure attempts to find any feasible $X$. If it succeeds, then $U = h(X)$ is an upper bound for the minima. The method attempts to reduce $U$ by transforming $h$ into constraint $h_{U-\eta}(Y) = h(Y) - (U - \eta) \leq 0$, where $\eta$ is a step-size. A feasibility problem is solved with added constraint $h_{U-\eta}(X) \leq 0$. Successes update $U$ and $X$ and double the size of $\eta$, while failures halve the size of $\eta$. The search halts when $\eta$ falls below threshold $\epsilon$. Algorithm 1 can efficiently solve for $A_s$ and $Q_s$ given hundreds of features.

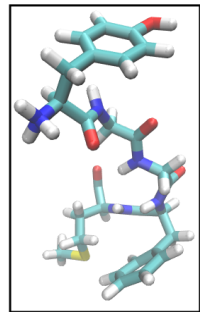

FIG. 2: Structure of met-enkephalin

Algorithm 2, restated here for completeness, minimizes convex function $\Phi(X)$ under the constraints that tr($X$) $\leq R$ over the course of $K$ descent steps[13]. The problems for $A_s$ and $Q_s$ each yield natural estimates for $R$. The primitive RESCALE applies a change of variable to transform the constraint tr($X$) $\leq R$ into the constraint tr($X$) $= 1$ by rescaling and adding a slack variable. The primitive APROX-EV computes the eigenvector associated with the largest eigenvalue of its argument. The Lanczos algorithm can perform this operation in time $O(d^2)$ per iteration[13]. The number of iterations required is $O(\log(N)\sqrt{C/\epsilon})$, where $C$ is any upper bound on the spectral norm of $X$.

#### 1. Numerical Issues

If the gradient at $X$ has largest eigenvalue near zero, the Lanczos algorithm can fail. In these cases, we shift the matrix upwards by a multiple of the identity[19]. If the Lanczos algorithm fails to converge even after shifting, we fall back to a slower but more stable LAPACK implementation of a divide-and-conquer eigenvalue algorithm.

### V. EXPERIMENTS

In this section, we learn metastable switching models from two MD simulation datasets. Our results indicate that metastable switching models can learn realistic models of protein dynamics.

### A. Met-enkephalin

Met-enkephalin is a small, naturally-occurring opioid peptide with five amino-acids and 75 atoms[14]. To study the dynamics of met-enkephalin, we used a collection of publicly available MD trajectories[20]. We superposed the raw trajectories upon a starting structure to eliminate rotational tumbling. We encoded the $(x, y, z)$ coordinates of all atoms in the superposed trajectories to obtain a time-series with $10^6$ data points, each with $d = 225$ features.

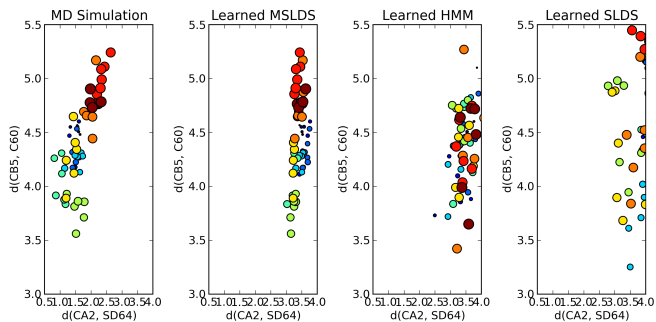

FIG. 3: Comparison of trajectories. Blue to red and small to large denote increasing time

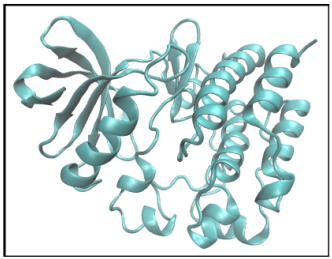

FIG. 4: Cartoon of Src-kinase

Figure 2 displays the structure of met-enkephalin, and Figure 3 compares trajectories from the original MD dataset (far-left) to trajectories sampled from a learned 2-state metastable switching linear dynamical system (MSLDS, center left), trajectories sampled from a learned 2-state HMM (center right), and trajectories sampled from a learned 2-state switching linear dynamical system without constraint $Q_s + A_s \Sigma_s A_s^T \preceq \Sigma_s$ (SLDS, right). The plots are projected onto chemically relevant order parameters for visualization. The scattered points shift in color from blue to red (and from small to large) as time increases. The MSLDS preserves the temporal coherence of the original data, while the HMM does not. The SLDS degenerates and does not capture the true distribution of the data.

### B. Src-Kinase

Protein kinases are enzymes critical to the cellular regulation network. The Src-Kinase is a member of this family with 262 amino acids. We obtained a dataset that samples $550\mu s$ of dynamics from this system for a total of 108GB of data[6]. To reduce the size of the feature space (the original system has over $4 \times 10^5$ atoms including waters), we selected the $(x, y, z)$ coordinates of 121 critical atoms for a total of $d = 363$ features. We then superposed upon a starting structure and subsampled every tenth data point from the featurized dataset, which reduced the data size to 1.8GB.

Figure 4 displays the Src-kinase enzyme. Figure 5 displays multiple sampled trajectories from a learned 3-state MSLDS model. The colors rise from blue to red as time increases. The model identifies three crucial regions previously discovered in the energy landscape[15]. More interestingly, the MSLDS model samples transition paths between these metastable regions, a feat not achievable with HMMs or MSMs. Such paths may be useful for the rational design of kinase inhibitors[15].

## VI. DISCUSSION AND CONCLUSION

This work presents an approach for learning protein dynamics from data and provides three major contributions: (1) The introduction of the metastable switching model, which contains physically inspired constraint $Q_s + A_s \Sigma_s A_s^T \preceq \Sigma_s$, and the derivation of associated inference and learning algorithms; (2) The extension of Frank-Wolfe algorithms to a broader class of convex programs; (3) The demonstration that metastable switching models can accurately learn the temporal dynamics of met-enkephalin and generate transition paths between metastable states of Src-kinase. Future work might apply metastable switching models to other physical datasets, such as temporal data from quantum-mechanical simulations.

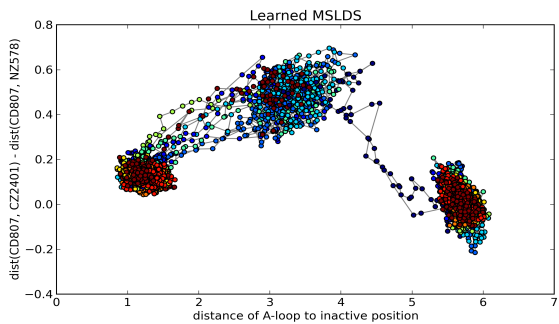

FIG. 5: Multiple metastable switching trajectories for Src-kinase sample transition paths. Blue to red denotes increasing time.

[1] Ken A Dill, Sarina Bromberg, Kaizhi Yue, Hue Sun Chan, Klaus M Fiebig, David P Yee, and Paul D Thomas. Principles of protein folding – a perspective from simple exact models. *Protein Science*, 4(4):561–602, 1995.
[2] Kyle A Beauchamp, Yu-Shan Lin, Rhiju Das, and Vijay S Pande. Are protein force fields getting better? a systematic benchmark on 524 diverse NMR measurements. *Journal of chemical theory and computation*, 8(4):1409–1414, 2012.

[3] David E Shaw et al. Anton, a special-purpose machine for molecular dynamics simulation. *Communications of the ACM*, 51(7):91–97, 2008.

[4] Alexandros Altis, Phuong H Nguyen, Rainer Hegger, and Gerhard Stock. Dihedral angle principal component analysis of molecular dynamics simulations. *The Journal of chemical physics*, 126(24):244111, 2007.

[5] Gregory R Bowman, Xuhui Huang, and Vijay S Pande. Using generalized ensemble simulations and Markov state models to identify conformational states. *Methods*, 49(2):197–201, 2009.

[6] Robert T McGibbon, Bharath Ramsundar, Mohammad M Sultan, Gert Kiss, and Vijay S Pande. Understanding protein dynamics with L1-regularized reversible hidden Markov models. *arXiv preprint arXiv:1405.1444*, 2014.

[7] Kai J Kohlhoff, Diwakar Shukla, Morgan Lawrenz, Gregory R Bowman, David E Konerding, Dan Belov, Russ B Altman, and Vijay S Pande. Cloud-based simulations on Google Exacycle reveal ligand modulation of GPCR activation pathways. *Nature chemistry*, 6(1):15–21, 2014.

[8] Michael L West and David P Fairlie. Targeting HIV-1 protease: a test of drug-design methodologies. *Trends in pharmacological sciences*, 16(2):67–75, 1995.

[9] Illia Horenko and Christof Schütte. Likelihood-based estimation of multidimensional Langevin models and its application to biomolecular dynamics. *Multiscale Modeling & Simulation*, 7(2):731–773, 2008.

[10] Kevin P Murphy. Switching kalman filters. Technical report, Citeseer, 1998.

[11] Sajid M Siddiqi, Byron Boots, and Geoffrey J Gordon. A constraint generation approach to learning stable linear dynamical systems. Technical report, DTIC Document, 2008.

[12] Christoph Helmberg, Franz Rendl, Robert J Vanderbei, and Henry Wolkowicz. An interior-point method for semidefinite programming. *SIAM Journal on Optimization*, 6(2):342–361, 1996.

[13] Elad Hazan. Sparse approximate solutions to semidefinite programs. In *LATIN 2008: Theoretical Informatics*, pages 306–316. Springer, 2008.

[14] Michael Comb, Peter H Seeburg, John Adelman, Lee Eiden, and Edward Herbert. Primary structure of the human met-and leu-enkephalin precursor and its mRNA. *Nature*, 295(5851):663–666, 1982.

[15] Diwakar Shukla, Yilin Meng, Benoît Roux, and Vijay S. Pande. Activation pathway of Src kinase reveals intermediate states as novel targets for drug design. *Nature Communications*, in press.

[16] Lawrence Rabiner. A tutorial on hidden markov models and selected applications in speech recognition. *Proceedings of the IEEE*, 77(2):257–286, 1989.

[17] Kenneth L Clarkson. Coresets, sparse greedy approximation, and the Frank-Wolfe algorithm. *ACM Transactions on Algorithms (TALG)*, 6(4):63, 2010.

[18] Stephen P Boyd and Lieven Vandenberghe. *Convex optimization*. Cambridge university press, 2004.

[19] Martin Jaggi, Marek Sulovsk, et al. A simple algorithm for nuclear norm regularized problems. In *Proceedings of the 27th International Conference on Machine Learning (ICML-10)*, pages 471–478, 2010.

[20] Met-enkephalin MD Trajectories. http://figshare.com/articles/Met_enkephalin_MD_Trajectories/1026324. Accessed: 2014-06-01.